\documentclass[letterpaper]{article}
\usepackage{spconf,graphicx, amsfonts}
\usepackage{epstopdf}
\usepackage[fleqn]{amsmath}
\usepackage{amssymb}
 \pdfoutput=1
\usepackage{algorithmicx}
\usepackage[ruled]{algorithm}
\usepackage{algpseudocode}
\alglanguage{pseudocode}
\usepackage[english]{babel}
\usepackage{fancyhdr}
\usepackage{threeparttable}
\usepackage{parskip}
\usepackage{cite}
\usepackage[bookmarks=false]{hyperref}

\usepackage{caption}
\usepackage{subfig}
\usepackage{float}
\usepackage{graphicx}
\usepackage{tablefootnote}
\usepackage{booktabs}
\usepackage{multirow}
\usepackage[T1]{fontenc}
\usepackage{pgf}
\usepackage[normalem]{ulem}
\usepackage[inline]{enumitem}
\usepackage[autoplay]{animate}

\usepackage{amsmath}
\usepackage{fixmath}
\usepackage{amsfonts}
\usepackage{xcolor}
\definecolor{colorsrc}{rgb}{0.36, 0.54, 0.66}

\definecolor{colorwnd2}{rgb}{0.91, 0.84, 0.42}
\definecolor{colorwnd}{rgb}{0.8, 0.0, 0.1}
\definecolor{colorfdd}{rgb}{0.44, 0.16, 0.39}
\definecolor{colorshi}{rgb}{0.55, 0.71, 0.0}
\definecolor{colornan}{rgb}{0.8, .33, 0}
\definecolor{colornan}{rgb}{0.72, 0.53, 0.04}
\definecolor{darkcyan}{rgb}{0.0, 0.55, 0.55}
\definecolor{colordfd}{rgb}{0.0, 0.2, 1.0}
\definecolor{colorlck}{rgb}{0.0, 0.42, 0.24}
\definecolor{pinegreen}{rgb}{0.0, 0.47, 0.44}

\def\bmt{\left[\begin{matrix}}

\def\emt{\end{matrix}\right]}

\def\bx{\mathbf{x}}
\def\ba{\mathbf{a}}

\def\bc{\mathbf{c}}
\def\bd{\mathbf{d}}

\def\br{\mathbf{r}}

\def\bu{\mathbf{u}}
\def\bw{\mathbf{w}}
\def\by{\mathbf{y}}
\def\bgamma{\pmb{\gamma}}

\def\brho{\pmb{\rho}}
\def\log{\text{log}}

\def\bA{\mathbf{A}}

\def\bC{\mathbf{C}}
\def\bD{\mathbf{D}}

\def\bL{\mathbf{L}}
\def\bl{\mathbf{l}}

\def\bI{\mathbf{I}}

\def\bn{\mathbf{n}}

\def\bz{\mathbf{z}}
\def\bv{\mathbf{v}}

\def\bzeros{\mathbf{0}}

\def\R{\mathbb{R}}

\newcommand{\norm}[1]{\left\|#1\right\|}

\makeatletter
\newsavebox\myboxA
\newsavebox\myboxB
\newlength\mylenA
\newcommand*\lbar[2][.75]{%
    \sbox{\myboxA}{$\m@th#2$}%
    \setbox\myboxB\null
    \ht\myboxB=\ht\myboxA%
    \dp\myboxB=\dp\myboxA%
    \wd\myboxB=#1\wd\myboxA
    \sbox\myboxB{$\m@th\overline{\copy\myboxB}$}
    \setlength\mylenA{\the\wd\myboxA}
    \addtolength\mylenA{-\the\wd\myboxB}%
    \ifdim\wd\myboxB<\wd\myboxA%
       \rlap{\hskip 0.5\mylenA\usebox\myboxB}{\usebox\myboxA}%
    \else
        \hskip -0.2\mylenA\rlap{\usebox\myboxA}{\hskip 0.2\mylenA\usebox\myboxB}%
    \fi}
\makeatother

\def\lbL{\lbar{\bL}}

\def\SS{\mathcal{S}}

\def\AS{\bD_{\SS}}
\def\DS{\bD_{\SS}}
\def\LS{\bL_{\SS}}

\def\xS{\bx^{\SS}}
\def\wS{\bw^{\SS}}
\def\rS{\br_{\SS}}
\def\UbarS{\lbar{U}_{\SS}}
\def\VbarS{\lbar{V}_{\SS}}
\def\sigmaS{\sigma_{\SS}}

\def\xShat{\bx^{\hat{\SS}}}
\def\Sandi{\SS \cup \{i\}}
\def\Ssubj{\SS \backslash \{j\}}

\title{\vspace{-.4in} ADAPTIVE MATCHING PURSUIT FOR SPARSE SIGNAL RECOVERY\vspace{-0.15in}}
\name{\vspace{-0.25in}Tiep H. Vu, Hojjat S. Mousavi, Vishal Monga \thanks{\noindent \hspace{-0.2in}This work has been supported partially by the Office of Naval Research (ONR) under Grant 0401531 UP719Z0 and NSF CAREER award to (V.M.).}}

\address{\vspace{-0.05in} {\small The Pennsylvania State University, University Park, PA}}

\begin{document}
\maketitle
\thispagestyle{fancy}
\pagestyle{empty}
\begin{abstract}
\vspace{0.1in}
\label{abstract}
Spike and Slab priors have been of much recent interest in signal processing as a means of inducing sparsity in Bayesian inference. Applications domains that benefit from the use of these priors include sparse recovery, regression and classification. It is well-known that solving for the sparse coefficient vector to maximize these priors results in a hard non-convex and mixed integer programming problem. Most existing solutions to this optimization problem either involve simplifying assumptions/relaxations or are computationally expensive. We propose a new greedy and adaptive matching pursuit (AMP) algorithm to directly solve this hard problem. Essentially, in each step of the algorithm, the set of active elements would be updated by either adding or removing one index, whichever results in better improvement. In addition, the intermediate steps of the algorithm are calculated via an inexpensive Cholesky decomposition which makes the algorithm much faster. Results on simulated data sets as well as real-world image recovery challenges confirm the benefits of the proposed AMP, particularly in providing a superior cost-quality trade-off over existing alternatives.

\end{abstract}


\vspace{-0.1in}
\section{Introduction}
\vspace{-0.1in}

\label{sec:intro}

Over the past decade, sparsity has become one of the most prevalent themes in signal
processing applications. In general, parsimony in signals describes the phenomenon where high dimensional data   can be expressed by only a few measurements. Sparse models assume that a signal can be efficiently represented as sparse linear combination of atoms in a given or learned dictionary \cite{Wright:SRC_PAMI2009,Sprechmann:CHI-LASSO_TSP2011}. 
The presence of sparsity in signals often enables us to provide efficient algorithms for extracting relevant information from the underlying data and is often  a natural assumption in inverse problems with variety of applications in image/signal classification \cite{Wright:SRC_PAMI2009,Srinivas:SSPIC_TIP2015}, dictionary learning \cite{Pourkamali:CompresiveKSVD_ICASSP2013, vu2015dfdl,vu2016tmi ,vu2016icip}, signal recovery \cite{Wright:SpaRSA_TSP2009, Tropp:OMP_InfoTheory2007, mousavi2015iterative}, image denoising and inpainting \cite{Elad:Sparsity_Denoise_2006TIP}, etc.

A sparse reconstruction algorithm aims to recover a
sparse signal $\bx \in \mathbb{R}^{p }$ from a set of fewer linear measurements $\by \in \mathbb{R}^{q}$ ($q\ll p$) according to: $\by = \bA \bx + \bn$, where $\bA \in \mathbb{R}^{q\times p}$ is the measurement matrix  and $\bn \in \mathbb{R}^{q}$ represents Gaussian noise.
Many solutions have been proposed for this problem and they include sparsity promoting optimization problems involving different regularizers such as $\ell_1$ or $\ell_0$ norms, greedy (e.g. matching pursuit) algorithms \cite{Tropp:OMP_InfoTheory2007, Mohimani:fast_l_0_TSP2009}, Bayesian-based methods \cite{JiAndCarin:BayesianCS_TSP2008, Lu:SparseCodeBayesPerspec_NeuralNetLearn2013} or general sparse approximation algorithms -- SpaRSA \cite{Wright:SpaRSA_TSP2009}, ADMM \cite{ Boyd:ADMM_MachineLearn2011}, etc. 
Many of these sparse recovery methods have shown that adding structural constraints and prior information to the frameworks have value in terms of representation purposes \cite{Sprechmann:CHI-LASSO_TSP2011, Srinivas:SSPIC_ICIP2013} and often leads to performance improvement.
Introducing priors for capturing sparsity as an example of Bayesian inference has shown to be effective for signal recovery \cite{Jenatton:StructVariableSelect_MAchineLearnResearch2011, Zou:VarSelecElasticNet_StatSociet2005}. Examples of such priors in statistics and signal processing are Laplacian \cite{Babacan_BayesianCSLaplacePriors_TIP2010}, generalized Pareto \cite{Cevher:SparseRecovGraphicalModel_SPMagaz2010}, Spike and Slab \cite{Mitchell:BayesVarSelectSpikeSlab_StatAssoc1988}, etc.  Amongst these priors, a well-suited sparsity promoting prior is Spike and Slab prior which is widely used in Bayesian inference \cite{Ishwaran_SpikeSlab_AnnStat2005,  Andersen:BayesianSpikeSlab_NIPS2014, Lazaro:SpikeSlabInferMultiTask_NIPS2011}. In fact, it is acknowledged that Spike and Slab prior is indeed the \emph{gold standard} for inducing sparsity in Bayesian inference \cite{Lazaro:SpikeSlabInferMultiTask_NIPS2011}.

In this paper in particular, we focus on Spike and Slab priors, introduced by \emph{Yen et al.}  \cite{Yen:MM_VariableSelectionSpikeSlab_Stat2011}, where every coefficient $x_i$ is modeled as a mixture of two densities as follows:
\begin{eqnarray*}
    x_i \sim (1-w_i) \mathbb{I}(x_i = 0) + w_i P_i(x_i). \label{Eq:GeneralSpikeSlab}
\end{eqnarray*}
$\mathbb{I}(\cdot)$ is the indicator function at zero (spike) and $P_i$ (slab) is a suitable prior distribution, e.g., Gaussian, for nonzero values of $x_i$. $w_i \in \{0,1\}$ controls the activeness of $x_i$. 

\noindent\textbf{Optimization Problem} (Hierarchical Bayesian Framework):
As suggested by \cite{Cevher_LearningCompressiblePriors_NIPS2009, Cevher:SparseRecovGraphicalModel_SPMagaz2010}, sparsity can be induced via a prior maximization procedure in a Bayesian setup. In this work, we employ Spike and Slab prior for inducing sparsity on $\bx$ and formulate a hierarchical Bayesian framework as in \cite{mousavi2015iterative}.

More precisely, the Bayesian formulation is as follows:
\begin{eqnarray*}
\footnotesize
    \by | \bA, \bx, \bgamma, \sigma^2   &\sim& \mathcal{N}(\bA\bx, \sigma^2 \bI), \\
\displaystyle   \bx | \bgamma, \lambda, \sigma^2   &\sim&  \prod_{i=1}^p \gamma_i \mathcal{N}(0, \sigma^2 \lambda^{-1}) + (1 - \gamma_i)\mathbb{I}(x_i = 0), \\
\bgamma &\sim&  \prod_{i=1}^p \text{Bernoulli}(\kappa_i),
\end{eqnarray*}
where $\mathcal{N}(.)$ represents the Gaussian distribution and $\bgamma$ is the indicator variable for vector $\bx$, i.e., $\gamma_i = 0$ if $x_i$ is zero, otherwise $\gamma_i=1$. The parameter $\kappa_i$ affects the sparsity level of the $\bx$ by separately controlling whether each indicator variable $\gamma_i$ is active or not.

\noindent The MAP estimation based on the above mentioned Bayesian framework leads to the optimization problem below \cite{mousavi2015iterative}:
 \begin{equation}
    \label{eqn:main}
     (\bx^*,\bgamma^* ) = \arg \min_{\bx, \bgamma} \norm{\by - \bA \bx}_2^2 + \lambda\norm{\bx}_2^2 + \sum_{i = 1}^{p} \rho_i \gamma_i,
 \end{equation}
and $\rho_i \triangleq \sigma^2 \log(\frac{2\pi \sigma^2 (1 - \kappa_i)^2}{\lambda \kappa_i^2})$. Note that from this definition, $\rho_i$ may be negative (if $\kappa_i$ is large enough).

This is a more general sparsity inducing optimization problem than the ones containing only $l_0$ or $l_1$ regularizers and has broad applicability in recovery and regression problems and is known to be a hard non-convex mixed integer programming. In \cite{Yen:MM_VariableSelectionSpikeSlab_Stat2011, Srinivas:SSPIC_TIP2015,Andersen:BayesianSpikeSlab_NIPS2014} simplifications are pursued by assuming a common {$\kappa$ ($\rho$)} for each coefficient, which reduces the last term in (\ref{eqn:main}) to {$\rho\|\mathbf{x}\|_{0}$}. Further, relaxation of the $\|\mathbf{x}\|_{0}$ to $\|\mathbf{x}\|_{1}$ leads to the well-known Elastic Net \cite{Zou:VarSelecElasticNet_StatSociet2005}. More recently an iterative refinement solution  is also proposed in \cite{mousavi2015iterative} which refines the solution at each step by considering a history of solutions at previous iterations. 
The {\bf Main contributions} of this paper are as follows: 1) To directly solve the hard non-convex problem in \eqref{eqn:main}, we propose an adaptive matching pursuit (AMP) procedure which is supported with theoretical analysis that formally argues the effectiveness as well as computational benefits of the algorithm. 2) The procedure can be slightly modified to solve problems with non-negativity constraints, which is often required in real-world applications. 3) We perform experimental validation on both simulated data and a practical image recovery problem, that reveals the merits of the proposed AMP; the practical findings also support the aforementioned analytical results.

\section{Adaptive Matching Pursuit} 
\label{sec:main_algorithm}
\vspace{-0.1in}
\label{sub:notation}

In this section, we propose a greedy solution for \eqref{eqn:main} by adding/removing elements to/from the support of $\bx$. First, let
$\bD = \bmt \bA\\\sqrt{\lambda} \bI \emt$ and $\bz = \bmt \by \\ \mathbf{0}\emt$ with $\bI \in \R^{p\times p}$ and $\mathbf{0} \in \R^{p\times 1}$ being the identity and zero matrices, we can rewrite (\ref{eqn:main}) as:
\vspace{-.1in}
\begin{equation}
\label{eqn:simplemain}
    (\bx^*, \bgamma^*) = \arg\min_{\bx, \bgamma} \norm{\bz - \bD\bx}_2^2 + \sum_{i=1}^p \rho_i \gamma_i.
\end{equation}
\noindent Note that we assume that each column of $\bA$ has norm 1, i.e. $\|\ba_i\|_2^2 = 1$ and subsequently, $\|\bd_i\|_2^2 = \|\ba_i\|_2^2 + \lambda = 1 + \lambda, ~\forall i = 1, \dots, p$.
It is also crucial to note that, if we know the true support of the signal, i.e. $\mathcal{S} = \{i: \gamma_i \neq 0 \}$, we can easily find the solution of (\ref{eqn:simplemain}) by calculating:
\begin{align}
      \label{eqn:xs}
    \xS = \arg\min_{\xS} \|\bz - \bD_{\SS}\xS\|_2^2 {~\Rightarrow \DS^T\DS\xS = \wS}\\
    \text{and~~} \rS = \bz - \DS\xS ~(\text{residual generated by}~ \SS), \nonumber
\end{align}
where $\wS$ and $\bD_{\SS}$ are a sub-vector of $\bw = \bD^T\by$ indexed by $\SS$ and a sub-matrix of $\bD$ formed by collecting its columns indexed by $\SS$, respectively. $\xS$ is the vector containing only active coefficients of $\bx$ indexed by $\SS$. 
The bijection $\SS \leftrightarrow \xS$ implies that solving (\ref{eqn:simplemain}) is equivalent to finding the active set $\SS$. This motivates us to utilize a greedy approach to find the support set $\SS$ and then infer the solution $\bx$. In particular, an Adaptive Matching Pursuit (AMP) is proposed to update the active set $\SS$ at each step by either absorbing one of the {\it unselected} indices into $\SS$ or removing one of active elements in $\SS$, whichever results in a larger decrease in the cost function.
Concretely, for a given $\SS$, we define:
\vspace{-0.07in}
\begin{equation}
    \label{eqn:deffS}
    \sigmaS = \sum_{i \in \SS}\rho_i \text{~~and~~} g(\SS) = \min_{\xS}\|\bz - \bD_{\SS}\xS\|_2^2 + \sigmaS.
\end{equation}
At each iteration, we  calculate two {\it "improvement"} values:
\begin{eqnarray}
    \label{eqn:defU}
    U_{\SS} &=& \min_{i\notin \SS}g(\SS \cup \{i\}) - g(\SS), \\
    V_{\SS} &=& \min_{j \in \SS}g(\SS \backslash \{j\}) - g(\SS).  \label{eqn:defB}
\end{eqnarray}
\emph{\eqref{eqn:defU} is the decrease in cost function if} \textbf{\textit{selecting}} \emph{one of unselected indices in the support and \eqref{eqn:defB} is the decrease if} \textbf{\textit{removing}} \emph{one already-selected index}.
If both $U_{\SS}$ and $V_{\SS}$ greater than or equal to $0$, we can stop the algorithm since no improvement is obtained. Otherwise, we compare $U_{\SS}$ and $V_{\SS}$ to update $\SS$ by absorbing $i$ (if $U_{\SS} < V_{\SS}$) or removing $j$ (if $U_{\SS} > V_{\SS}$). This procedure guarantees that the cost function decreases after each iteration, and then, the algorithm will eventually stop after finite iterations.

\par Nevertheless, the cost of calculating $g(\SS \cup \{i\})$ and $g(\SS \backslash \{j\})$ is extremely high and this idea becomes hardly practical. In order to significantly reduce the  computation cost while keeping our algorithm close to the idea above, instead of exactly calculating $U_{\SS}$ and $V_{\SS}$, we aim to calculate their competent upper bounds $\lbar{U}_{\SS}$ and $\lbar{V}_{\SS}$. The decision is then made based on these approximated values to obtain $\SS^{\text{new}}$. After that, $\bx^{\SS^{\text{new}}}$ and $\br_{\SS^{\text{new}}}$ are calculated {\it precisely} before moving to the new iteration.
The following lemmas support AMP in initializing $\SS$ and choosing $\lbar{U}_{\SS}$ and $\lbar{V}_{\SS}$.

\noindent \textbf{Lemma 1:} If $\rho_i < 0$, then $i \in \hat{\SS}$ -- the optimal active set.
\par
\noindent \textit{Proof:} 
Suppose that $i \notin \hat{\SS}$. Let $\br_{\hat{\SS}} = \bz - \bD_{\hat{\SS}}\xShat$, we have:
\noindent \hspace{-.2in}
\begin{align}
    \nonumber
   \hspace{-0.35in} g(\hat{\SS} \cup \{i\})& \stackrel{\text{by (\ref{eqn:deffS})}}{\leq}
    \|\br_{\hat{\SS}} - x_i\bd_i\|_2^2 + \sigma_{\hat{\SS}} + \rho_i\\
    \nonumber
    &= g(\hat{\SS}) + \bd_i^T\bd_i x_i^2 - 2\br_{\hat{\SS}}^T\bd_i x_i + \rho_i\\
\label{eqn:lem1}
    &= g(\hat{\SS}) + (1 + \lambda) x_i^2 - 2\br_{\hat{\SS}}^T\bd_i x_i + \rho_i.
\end{align}
\par{}
Let $h(x) = (1 + \lambda) x^2 - 2\br_{\hat{\SS}}^T\bd_i x + \rho_i$, we observe that $h(x)$ is continuous and
     $\lim_{x \rightarrow \infty} h(x) = +\infty ~\text{and}~ h(0) = \rho_i < 0$.
Then, there exists $\lbar{x}_i \neq 0$ such that $\rho_i < h(\lbar{x}_i) < 0.$ Combining this with (\ref{eqn:lem1}), we have:
\begin{equation*}
    g(\hat{\SS} \cup \{i\}) \leq g(\hat{\SS}) + h(\lbar{x}_i) < g(\hat{\SS}).
\end{equation*}
In other words, $\hat{\SS} \cup \{i\}$ generates a lower cost than $\hat{\SS}$ does, which contradicts the assumption that $\hat{\SS}$ is the optimal solution. The contradiction suggests that $i$ must be in $\hat{\SS}$. \hfill $\blacksquare$

\noindent By using Lemma 1, we can initialize $\SS_0 = \{i: \rho_i < 0\}$.
\begin{equation}
\label{eqn:lem2}
    \textbf{Lemma 2:~~~} U_{\SS} \leq \min_{i \notin \SS}\left\{ \rho_i - \frac{(\rS^T\bd_i)^2}{1 + \lambda}\right\} \triangleq \UbarS.~~~~~~~~~~~~~~~
\end{equation}
\noindent\textit{Proof:} Since (\ref{eqn:lem1}) holds for every $x_i$, we have:
\begin{align}
\nonumber
    g(\SS \cup \{i\}) \leq& g(\SS) + \min_{x_i}\left\{ \rho_i + (1+\lambda)x_i^2 - 2\rS^T\bd_ix_i\right\}\\
    \label{eqn:lem22}
    =&g(\SS) + \rho_i - \frac{(\rS^T\bd_i)^2}{1 + \lambda}.
\end{align}
Since (\ref{eqn:lem22}) holds for every $i \notin \SS$, inequality (\ref{eqn:lem2}) is true.
\hfill $\blacksquare$

\noindent\textbf{Lemma 3:}
\begin{equation}
    \label{eqn:mainlem3}
    V_{\SS} \leq \min_{j \in \SS}\left\{ (1 + \lambda)(x^{\SS}_j)^2 + 2\bd_j^T\rS x^{\SS}_j - \rho_j\right\} \triangleq \VbarS.
\end{equation}
where $x^{\SS}_j$ is the element of $\xS$ corresponding to index $j \in \SS$.
\par \noindent\textit{Proof:} For any $j \in \SS$,
\begin{align}
\hspace{-.3in}
\nonumber
    g(\SS) =&~ \|\rS + \bd_j x^{\SS}_j - \bd_j x^{\SS}_j\|_2^2  + \sigma_{\SS \backslash \{j\}} + \rho_j \\
    \nonumber
    =&~ \|\rS + \bd_jx^{\SS}_j\|_2^2 +  \sigma_{\SS \backslash \{j\}} + \\
    \nonumber
    & ~+(1 + \lambda) (x_j^{\SS})^2 + \rho_j - 2\bd_j^T(\rS + \bd_jx^{\SS}_j)x_j^{\SS} \\
    \label{eqn:lem3}
    \geq
    &~ g(\SS \backslash \{j\}) - (1 + \lambda) (x_j^{\SS})^2 - 2\bd_j^T\rS x_j^{\SS} + \rho_j.
\end{align}
Combining (\ref{eqn:lem3}) and (\ref{eqn:defB}) we can conclude (\ref{eqn:mainlem3}). \hfill $\blacksquare$

From Lemma 2 and Lemma 3, we can consider $\lbar{U}_{\SS}$ and $\lbar{V}_{\SS}$ as good approximations of $U_{\SS}$ and $V_{\SS}$ and update the support set $\SS$ based on those. It is worth to note here that calculating $\lbar{U}_{\SS}$ and $\lbar{V}_{\SS}$ requires low computations.

\par {After obtaining updated $\SS$, we have to find $\xS$ using \eqref{eqn:xs} where we present a computationally cheap solution for that.
Let $\DS^T\DS = \LS\LS^T$ be the Cholesky decomposition of the symmetric positive definite matrix $\DS^T\DS$ where $\LS$ is a low triangular matrix.
Then, we can rewrite (\ref{eqn:xs}) as $\LS\LS^T\xS = \wS$. Subsequently, $\xS$ can be updated by solving two equations $\LS \bu = \wS$ and $\LS^T\xS = \bu$, consecutively. These two equations are simply solved by forward substitution and backward substitution, respectively, thanks to the triangularity of $\LS$. The remaining task now is that given $\LS$, how do we effectively obtain $\bL_{\Sandi}$ and $\bL_{\Ssubj}$? While the $\bL_{\Sandi}$ update procedure is quite classical, as far as we know, the $\bL_{\Ssubj}$ update procedure has not been widely addressed. Details of these two procedures are given below:}


\noindent 1. {\it Given $\LS$, calculate $\bL_{\Sandi}$}. This problem has been tackled in \cite{Rubinstein2008}.
 Specifically, by writing $\bD_{\Sandi} = \bmt \DS & \bd_i\emt$, we have:
 \begin{eqnarray*}
 \nonumber
     \bL_{\Sandi}\bL_{\Sandi}^T &=& \bD_{\Sandi}^T\bD_{\Sandi} \\
     &=& \bmt \DS^T\DS &\DS^T\bd_i \\ \bd_i^T\DS &\bd_i^T\bd_i \emt
     = \bmt \LS\LS^T & \DS^T\bd_i \\ \bd_i^T\DS & 1 + \lambda\emt.
 \end{eqnarray*}
If $\bv$ is the solution of $\LS \bv = \DS^T\bd_i$ via forward substitution, then we can easily infer:
\begin{equation}
\label{eqn:Linsert}
    \bL_{\Sandi} = \bmt \LS & \mathbf{0} \\ \bv^T & \sqrt{1 + \lambda - \bv^T\bv} \emt.
\end{equation}
\noindent 2. {\it Given $\LS$, calculate $\bL_{\Ssubj}$}. Suppose that: $\DS^T\DS = $
{\footnotesize
\begin{equation*}
   =\underbrace{\bmt \bL_{11} & \bzeros & \bzeros \\
                                \bl_{21}^T & l_{22} & \bzeros \\
                                \bL_{31} & \bl_{32} & \bL_{33} \emt}_{\bL_{\SS}}
                \underbrace{\bmt \bL_{11}^T &\bl_{21} & \bL_{31}^T \\
                                \bzeros & l_{22} & \bl_{32}^T \\
                                \bzeros & \bzeros & \bL_{33}^T\emt}_{\bL_{\SS}^T}
    = \bmt \bC_{11} & \bc_{12} &\bC_{13} \\
                    \bc_{12}^T & c_{22} & \bc_{32}^T \\
                    \bC_{31} & \bc_{32} & \bC_{33} \emt, %
\end{equation*}}%
where $[\bc_{12}^T, c_{22}, \bc_{32}^T]$ and $[\bc_{12}^T, c_{22}, \bc_{32}^T]^T$ are the row and column corresponding to the removed index $j$. By writing:
\vspace{-.1in}
\begin{equation*}
    \bD_{\Ssubj}^T\bD_{\Ssubj} = \bmt \bC_{11} & \bC_{13} \\ \bC_{31}& \bC_{33} \emt
    = \underbrace{\bmt \lbL_{11} & \bzeros \\
                \lbL_{31} & \lbL_{33}
            \emt}_{\bL_{\Ssubj}}
        \underbrace{\bmt \lbL_{11}^T & \lbL_{31}^T \\
                    \bzeros & \lbL_{33}^T
                \emt}_{\bL_{\Ssubj}^T},
\end{equation*}
we obtain:
\hspace{-.2in}
\begin{eqnarray}
    \bC_{11} &=& \bL_{11}\bL_{11}^T = \lbL_{11}\lbL_{11}^T  \Rightarrow \lbL_{11} = \bL_{11} \label{eqn:C11},\\
    \bC_{13} &=& \bL_{11}\bL_{31}^T = \lbL_{11} \lbL_{31}^T \Rightarrow \lbL_{31} = \bL_{31}, \nonumber\\
    \nonumber
    \bC_{33} &=& \bL_{31}\bL_{31}^T + \bl_{32}\bl_{32}^T + \bL_{33}\bL_{33}^T = \lbL_{31}\lbL_{31}^T + \lbL_{33}\lbL_{33}^T, \\
    &\Rightarrow& \lbL_{33}\lbL_{33}^T = \bL_{33}\bL_{33}^T + \bl_{32}\bl_{32}^T.\nonumber
\end{eqnarray}
Note that since Cholesky decomposition is unique we can obtain \eqref{eqn:C11}. Also since $\bl_{32}$ is a vector, $ \bl_{32}\bl_{32}^T$ is a rank-one matrix. Therefore, $\lbL_{33}$ can be obtained quickly from $\bL_{33}$ and $\bl_{32}$ using rank one update for the Cholesky decomposition \cite{seeger2004low} and we obtain:
\begin{equation}
\label{eqn:Lremove}
    \bL_{\Ssubj} = \bmt \bL_{11} & \bzeros \\
                        \bL_{31} & \lbL_{33}  \emt.
\end{equation}
Altogether, the AMP algorithm is presented in Algorithm \ref{alg:AMP}.
\begin{algorithm}[t]
\small
    \caption{AMP algorithm for solving problem (\ref{eqn:main})}\label{alg:AMP}
    \begin{algorithmic}
    \Function {$(\bx^*, \bgamma^*)$ = AMP}{$\by, \bA, \lambda, \brho$}.
    \State 1. $\bD = \bmt \bA\\\sqrt{\lambda} \bI \emt$ and $\bz = \bmt \by \\ \mathbf{0}\emt$
    \State 2. Initialize $\SS = \{i: \rho_i < 0\}$ and $\LS$: $\LS\LS^T = \DS^T\DS$.
    \State {\it \% if $\SS = \emptyset$, then $\bL = [~]$.}
    \While{ true}
    \State 3. Solve $\bu: \LS \bu = \wS \text{~~then solve ~~}\xS:   \LS^T\xS = \bu.$
    \State 4. Update residual: $\rS = \bz - \AS\xS.$
        \State 5. Calculate:
        \begin{math}
            \displaystyle
            [\UbarS, i] = \min_{i \notin \SS}\left\{ \rho_i - \frac{(\rS^T\bd_i)^2}{1 + \lambda}\right\}.
        \end{math}
        \State 6. Calculate $\VbarS$:
        \begin{equation*}
            [\VbarS, j] = \min_{j \in \SS}\left\{ (1 + \lambda)(x^{\SS}_j)^2 + 2\bd_j^T\rS x_j - \rho_j\right\}.
        \end{equation*}
        \State 7. Decide
        \If {$\min(\UbarS, \VbarS) \geq 0$}            break the while loop.
        \ElsIf {$\UbarS < \VbarS$}
            \State Insert index: $\SS = \SS\cup \{i\}$ and update $\LS$ by (\ref{eqn:Linsert}).
        \Else ~~Remove index: $\SS = \SS \backslash \{j\}$ and update $\LS$ by (\ref{eqn:Lremove}).
        \EndIf
    \EndWhile
    \State 8. {\bf OUTPUT:} $\SS \Rightarrow \bgamma^*;~~\xS \Rightarrow \bx^*$
    \EndFunction
    \end{algorithmic}
\end{algorithm}
\begin{figure*}[!ht]
\centering
  \includegraphics[width = .95\textwidth]{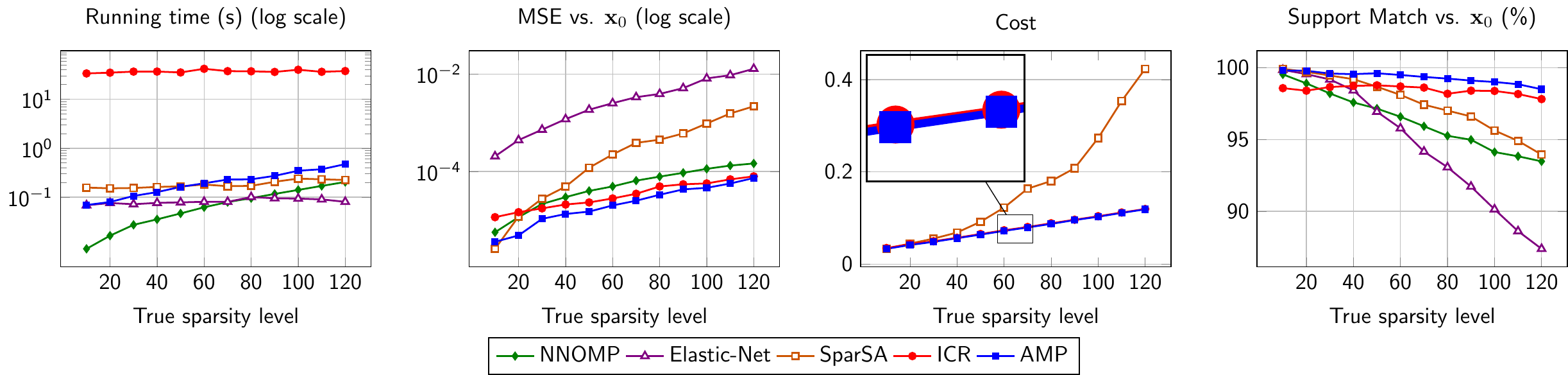}
\caption{Comparison of (from left to right): running time, mean squared error (MSE), cost function and  Support Match (SM) obtained by each method versus sparsity level of $\bx_0$ with non-negativity constraint ($p = 512, q = 256$).}
\label{fig:result2}
\vspace{-.1in}
\end{figure*}

\noindent\textbf{Remark:} Problem (\ref{eqn:main}) with non-negativity constraint can also be solved by slightly modifying Step 3 and Step 5 of Algorithm \ref{alg:AMP}. In Step 3, $\xS$ is instead solved via:
\begin{math}
\displaystyle
    \xS = \arg\min_{\xS \succeq 0} \|\bz - \DS\xS\|_2^2
\end{math}; while in Step 5, the alternative is: $[\UbarS, i] = \min_{i \notin \SS}\left\{ \rho_i - (\max\{\rS^T\bd_i, 0\})^2/(1 + \lambda)\right\}$.
The non-negative quadratic programming in Step 3 can be solved by an ADMM \cite{Boyd:ADMM_MachineLearn2011} procedure as proposed in \cite{boley2013local} which can also benefit from our Cholesky decomposition to make it computationally cheap\footnote{{Details of this procedure will be discussed in our future work.}}.

\begin{table}[]
\centering
\caption{Comparison of methods for $p = 512$, $q = 256$.}
\label{tab: result1}
\small{\begin{tabular}{|l||c|c|c|c||c|}
\hline
Average  & CoSaMP  & E-NET       & SpaRSA & ICR    & {\bf AMP}    \\ \hline \hline
Time (s) & 1.1E0  & {\bf 7.9E-2} & 2.2E-1 & 2.9E1  & 1.6E-1 \\ \hline
MSE      & 3.0E-4 & 4.5E-3       & 5.4E-4 & 3.1E-3 & {\bf 6.1E-5} \\ \hline
Cost     & -      & -            & 1.6E-1 & 2.1E-1 & {\bf 9.5E-2} \\ \hline
SM (\%)  & 93.52  & 85.55        & 93.87  & 83.79  & {\bf 97.32}  \\ \hline
\end{tabular}}
\end{table}

\section{Experimental results} 

\label{sec:experimental_results}{}
To illustrate the effectiveness of our AMP algorithm, we apply it to sparse recovery problems in two different scenarios: i) simulated data and ii) a real-world image recovery problem. Comparisons are made against state-of-the-art alternatives: 1) {CoSaMP \cite{needell2009cosamp} / NNOMP \cite{bruckstein2008uniqueness} \footnote{CoSaMP for unconstrained OMP, NNOMP for non-negative OMP.}}
 ; 2) Elastic Net (E-NET) \cite{Zou:VarSelecElasticNet_StatSociet2005} 3) SpaRSA \cite{Wright:SpaRSA_TSP2009} and 4) ICR \cite{mousavi2015iterative}.
\par \noindent  \textit{Simulated data:}
We set up a typical experiment for sparse recovery as in \cite{beck2009fast,Yen:MM_VariableSelectionSpikeSlab_Stat2011} with a randomly generated Gaussian matrix $\bA \in \R^{256 \times 512}$ and a sparse vector $\bx_0 \in \R^{512}$ with 100 non-zeros. Based on $\bA$ and $\bx_0$, we form the observation vector $\by \in \R^{256}$ according to the additive noise model with $\sigma = 0.01$: $\by = \bA\bx_0 + \bn$. Table \ref{tab: result1} reports the experimental results (averages of 100 trials with different $\bA, \bx_0$ and $\bn$) for this problem. As can be seen from Table \ref{tab: result1}, AMP outperforms the competing methods in many different aspects. {AMP runs almost two hundred times faster than ICR and seven times faster than CoSaMP}. Results in the second row of Table \ref{tab: result1} reveal that the AMP solution is the closest to the ``ground truth'' $\bx_0$ in terms of mean square error (MSE). In terms of the cost function, we compare AMP with two other methods, ICR and SpaRSA, which solve the same optimization problem (\ref{eqn:main}). The third row confirms that AMP achieves the lowest cost function (by more than one order of magnitude). Finally, we use ``support match'' (SM) to measure how much the support of each solution matches to that of $\bx_0$. It is evident that AMP provides significantly higher support match (SM = 97.21\%) than other state-of-the-art methods.
Next, in order to see how each method performs in the presence of non-negativity constraints, we perform one more experiment with sparse data that is naturally non-negative and vary the sparsity level of $\bx_0$ from 10 to 120 and compare the running time, MSE, optimal cost function and SM of different methods. Results are shown in Figure \ref{fig:result2}. Similar trends can be seen in this figure where {AMP requires less running time than ICR does} while it consistently outperforms others in the remaining aspects. It is worth to mention that AMP and ICR obtain almost the identical cost which is better than what SpaRSA achieves.
\par
\begin{figure}
\centering
  \includegraphics[width = .49\textwidth]{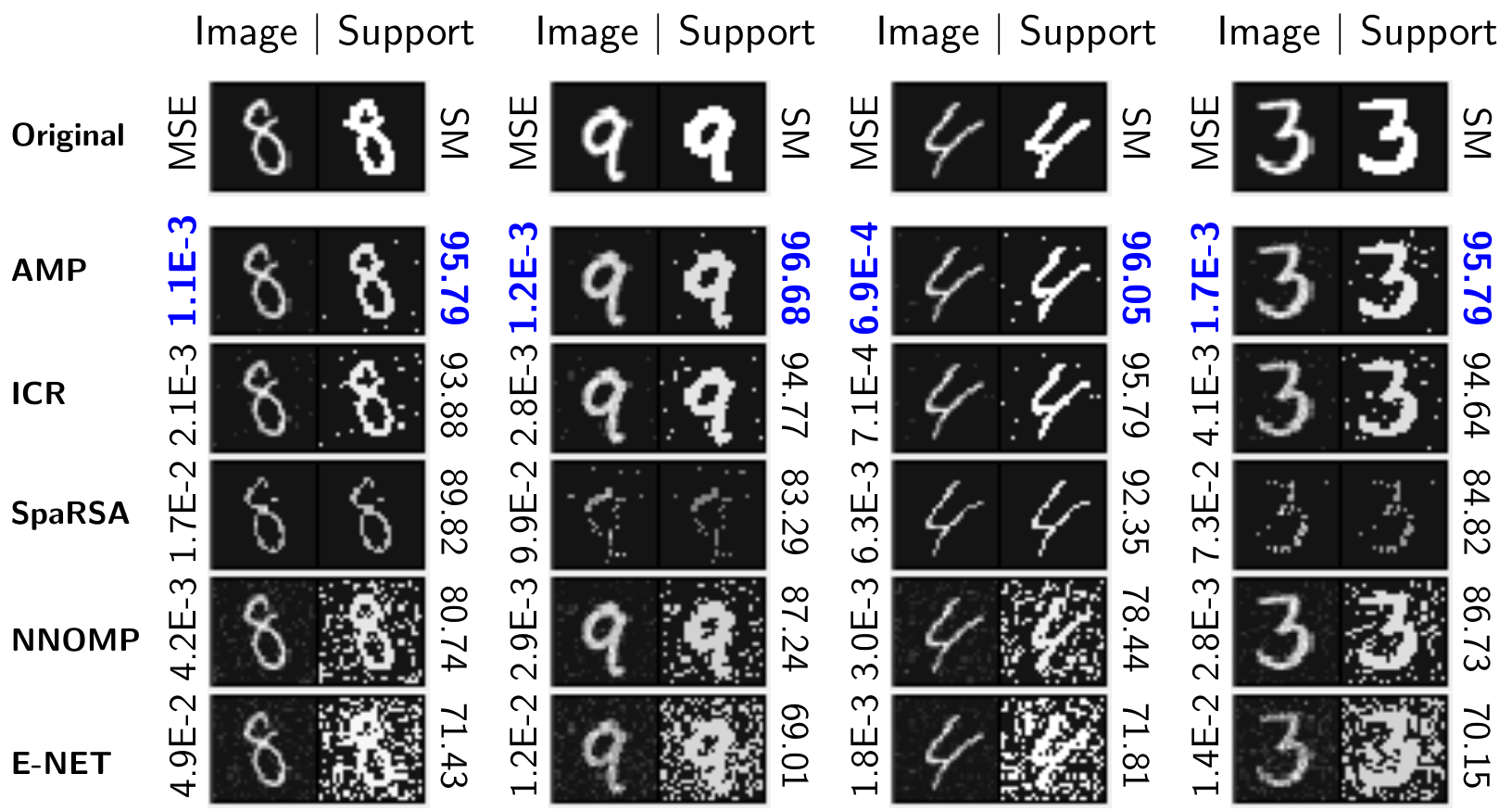}
  \caption{Examples of reconstructed images from MNIST dataset using different methods. In each pair of images: left is the original image $\bx_0$ (for the first row) or the reconstructed image (for other rows), right  is its ``support'' image. Numbers on the left are MSEs and on the right are Support Matches.}
\label{fig:mnist}
\end{figure}

 \noindent \textit{Image recovery:} Next, we apply the different sparse recovery algorithms to real data for image reconstruction. We work with the well-known handwritten digit images MNIST \cite{MINIST}. The dataset contains 60,000 digit images (0 to 9) of size $28\times 28$. Since most of pixels in each image are inactive (0), each image is naturally sparse. The experiment is set up such that a vectorized sparse signal $\bx \in \R^{784 \times 1}$ is to be reconstructed from a smaller set of random measurements $\by$. We randomly generate a Gaussian matrix $\bA \in \R^{350 \times 784}$. For each signal, we add a Gaussian noise $\bn$ with $\sigma = 0.03$. 
Recovered images and their supports are demonstrated in Figure \ref{fig:mnist} along with the corresponding MSE  and Support Match. 
Clearly, AMP and ICR outperform other methods with slightly better but much faster results provided by AMP.
\newpage
\pagebreak
\bibliographystyle{IEEEbib}
{\small\bibliography{ICASS_2017_arXiv}}

\begin{thebibliography}{10}

\bibitem{Wright:SRC_PAMI2009}
J.~Wright, Allen~Y Yang, Arvind Ganesh, Shankar~S Sastry, and Yi~Ma,
\newblock ``Robust face recognition via sparse representation,''
\newblock {\em IEEE Trans.\ on Pattern Analysis and Machine Int.}, vol. 31, no.
  2, pp. 210--227, 2009.

\bibitem{Sprechmann:CHI-LASSO_TSP2011}
P.~Sprechmann, I.~Ramirez, G.~Sapiro, and Y.~C. Eldar,
\newblock ``C-{H}i{L}{A}{S}{S}{O}: A collaborative hierarchical sparse modeling
  framework,''
\newblock {\em IEEE Trans.\ on Signal Processing}, vol. 59, no. 9, pp.
  4183--4198, 2011.

\bibitem{Srinivas:SSPIC_TIP2015}
U.~Srinivas, Y.~Suo, Minh Dao, V.~Monga, and T.~D Tran,
\newblock ``Structured sparse priors for image classification,''
\newblock {\em IEEE Transactions on Image Processing}, vol. 24, no. 6, pp.
  1763--1776, 2015.

\bibitem{Pourkamali:CompresiveKSVD_ICASSP2013}
F.~Pourkamali Anaraki and S.~M. Hughes,
\newblock ``Compressive {K}-{S}{V}{D},''
\newblock in {\em Proc.\ IEEE Int.\ on Conf.\ Acoustics, Speech, and Signal
  Processing}. IEEE, 2013, pp. 5469--5473.

\bibitem{vu2015dfdl}
T.~H. Vu, H.~S. Mousavi, V.~Monga, UK~Rao, and G.~Rao,
\newblock ``{D}{F}{D}{L}: Discriminative feature-oriented dictionary learning
  for histopathological image classification,''
\newblock {\em Proc. IEEE International Symposium on Biomedical Imaging}, pp.
  990--994, 2015.

\bibitem{vu2016tmi}
T.~H. Vu, H.~S. Mousavi, V.~Monga, UK~Rao, and G.~Rao,
\newblock ``Histopathological image classification using discriminative
  feature-oriented dictionary learning,''
\newblock {\em IEEE Transactions on Medical Imaging}, vol. 35, no. 3, pp.
  738--751, March, 2016.

\bibitem{vu2016icip}
T.~H. Vu and V.~Monga,
\newblock ``Learning a low-rank shared dictionary for object classification,''
\newblock {\em Proc.\ IEEE Conf.\ on Image Processing}, pp. 4428--4432.

\bibitem{Wright:SpaRSA_TSP2009}
S.~J. Wright, R.~D. Nowak, and M.~AT Figueiredo,
\newblock ``Sparse reconstruction by separable approximation,''
\newblock {\em IEEE Trans.\ on Signal Processing}, vol. 57, no. 7, pp.
  2479--2493, 2009.

\bibitem{Tropp:OMP_InfoTheory2007}
J.~A. Tropp and A.~C. Gilbert,
\newblock ``Signal recovery from random measurements via orthogonal matching
  pursuit,''
\newblock {\em IEEE Trans.\ on Info.\ Theory}, vol. 53, no. 12, pp. 4655--4666,
  2007.

\bibitem{mousavi2015iterative}
H.~S. Mousavi, V.~Monga, and T.~D. Tran,
\newblock ``Iterative convex refinement for sparse recovery,''
\newblock {\em IEEE Signal Processing Letters}, vol. 22, no. 11, pp.
  1903--1907, 2015.

\bibitem{Elad:Sparsity_Denoise_2006TIP}
M.~Elad and M.~Aharon,
\newblock ``Image denoising via sparse and redundant representations over
  learned dictionaries,''
\newblock {\em IEEE Trans.\ on Im. Processing}, vol. 15, no. 12, pp.
  3736--3745, 2006.

\bibitem{Mohimani:fast_l_0_TSP2009}
H.~Mohimani, M.~Babaie-Zadeh, and C.~Jutten,
\newblock ``A fast approach for overcomplete sparse decomposition based on
  smoothed norm,''
\newblock {\em IEEE Trans.\ on Signal Processing}, vol. 57, no. 1, pp.
  289--301, 2009.

\bibitem{JiAndCarin:BayesianCS_TSP2008}
S.~Ji, Y.~Xue, and L.~Carin,
\newblock ``Bayesian compressive sensing,''
\newblock {\em IEEE Trans.\ on Signal Processing}, vol. 56, no. 6, pp.
  2346--2356, 2008.

\bibitem{Lu:SparseCodeBayesPerspec_NeuralNetLearn2013}
X.~Lu, Y.~Wang, and Y.~Yuan,
\newblock ``Sparse coding from a {B}ayesian perspective,''
\newblock {\em Neural Networks and Learning Systems, IEEE Transactions on},
  vol. 24, no. 6, pp. 929--939, 2013.

\bibitem{Boyd:ADMM_MachineLearn2011}
S.~Boyd, N.~Parikh, E.~Chu, B.~Peleato, and J.~Eckstein,
\newblock ``Distributed optimization and statistical learning via the
  alternating direction method of multipliers,''
\newblock {\em Foundations and Trends{\textregistered} in Machine Learning},
  vol. 3, no. 1, pp. 1--122, 2011.

\bibitem{Srinivas:SSPIC_ICIP2013}
U.~Srinivas, Y.~Suo, Minh Dao, V.~Monga, and T.~D Tran,
\newblock ``Structured sparse priors for image classification.,''
\newblock in {\em Proc.\ IEEE Conf.\ on Image Processing}, 2013, pp.
  3211--3215.

\bibitem{Jenatton:StructVariableSelect_MAchineLearnResearch2011}
R.~Jenatton, J-Y Audibert, and F.~Bach,
\newblock ``Structured variable selection with sparsity-inducing norms,''
\newblock {\em The Journal of Machine Learning Research}, vol. 12, pp.
  2777--2824, 2011.

\bibitem{Zou:VarSelecElasticNet_StatSociet2005}
H.~Zou and T.~Hastie,
\newblock ``Regularization and variable selection via the elastic net,''
\newblock {\em Journal of the Royal Stat. Society: Series B (Stat.
  Methodology)}, vol. 67, no. 2, pp. 301--320, 2005.

\bibitem{Babacan_BayesianCSLaplacePriors_TIP2010}
S.D. Babacan, R.~Molina, and A.K. Katsaggelos,
\newblock ``Bayesian compressive sensing using laplace priors,''
\newblock {\em IEEE Trans.\ on Image Processing}, vol. 19, no. 1, pp. 53--63,
  2010.

\bibitem{Cevher:SparseRecovGraphicalModel_SPMagaz2010}
V.~Cevher, P.~Indyk, L.~Carin, and R.~G. Baraniuk,
\newblock ``Sparse signal recovery and acquisition with graphical models,''
\newblock {\em Signal Processing Magazine, IEEE}, vol. 27, no. 6, pp. 92--103,
  2010.

\bibitem{Mitchell:BayesVarSelectSpikeSlab_StatAssoc1988}
T.~J. Mitchell and J.~J. Beauchamp,
\newblock ``Bayesian variable selection in linear regression,''
\newblock {\em Journal of the American Statistical Association}, vol. 83, no.
  404, pp. 1023--1032, 1988.

\bibitem{Ishwaran_SpikeSlab_AnnStat2005}
H.~Ishwaran and J.~S. Rao,
\newblock ``{S}pike and {S}lab variable selection: frequentist and {B}ayesian
  strategies,''
\newblock {\em Annals of Statistics}.

\bibitem{Andersen:BayesianSpikeSlab_NIPS2014}
M.~R. Andersen, O.~Winther, and L.~K. Hansen,
\newblock ``Bayesian inference for structured {S}pike and {S}lab priors,''
\newblock in {\em Advances in Neural Information Processing Systems}, 2014, pp.
  1745--1753.

\bibitem{Lazaro:SpikeSlabInferMultiTask_NIPS2011}
M.~L{\'a}zaro-gredilla and M.~K. Titsias,
\newblock ``{S}pike and {S}lab variational inference for multi-task and
  multiple kernel learning,''
\newblock in {\em Advances in neural information processing systems}, 2011, pp.
  2339--2347.

\bibitem{Yen:MM_VariableSelectionSpikeSlab_Stat2011}
T.~J. Yen et~al.,
\newblock ``A majorization--minimization approach to variable selection using
  {S}pike and {S}lab priors,''
\newblock {\em The Annals of Statistics}, vol. 39, no. 3, pp. 1748--1775, 2011.

\bibitem{Cevher_LearningCompressiblePriors_NIPS2009}
V.~Cevher,
\newblock ``Learning with compressible priors,''
\newblock in {\em Advances in Neural Information Processing Systems}, 2009, pp.
  261--269.

\bibitem{Rubinstein2008}
R.~Rubinstein, M.~Zibulevsky, and M.~Elad,
\newblock ``Efficient implementation of the {K}-{S}{V}{D} algorithm using batch
  orthogonal matching pursuit,''
\newblock {\em CS Technion}, vol. 40, no. 8, pp. 1--15, 2008.

\bibitem{seeger2004low}
M.~Seeger,
\newblock ``Low rank updates for the {C}holesky decomposition,''
\newblock Tech. {R}ep., 2004.

\bibitem{boley2013local}
D.~Boley,
\newblock ``Local linear convergence of the alternating direction method of
  multipliers on quadratic or linear programs,''
\newblock {\em SIAM Journal on Optimization}, vol. 23, no. 4, pp. 2183--2207,
  2013.

\bibitem{needell2009cosamp}
D.~Needell and J.~A. Tropp,
\newblock ``Co{S}a{M}{P}: Iterative signal recovery from incomplete and
  inaccurate samples,''
\newblock {\em Applied and Computational Harmonic Analysis}, vol. 26, no. 3,
  pp. 301--321, 2009.

\bibitem{bruckstein2008uniqueness}
A.~M Bruckstein, M.~Elad, and M.~Zibulevsky,
\newblock ``On the uniqueness of nonnegative sparse solutions to
  underdetermined systems of equations,''
\newblock {\em IEEE Transactions on Information Theory}, vol. 54, no. 11, pp.
  4813--4820, 2008.

\bibitem{beck2009fast}
A.~Beck and M.~Teboulle,
\newblock ``A fast iterative shrinkage-thresholding algorithm for linear
  inverse problems,''
\newblock {\em SIAM Journal on Imaging Sciences}, vol. 2, no. 1, pp. 183--202,
  2009.

\bibitem{MINIST}
Y.~LeCun, C.~Cortes, and Burges C.~J.C.,
\newblock ``{M}{N}{I}{S}{T} dataset,'' \url{http://yann.lecun.com/exdb/mnist/},
\newblock Accessed: 2016-08-30.

\end{thebibliography}
\end{document}